\documentclass[letterpaper]{article}
\usepackage[preprint]{aaai2027}
\usepackage[hyphens]{url}
\usepackage{graphicx}
\usepackage{natbib}
\usepackage{caption}
\usepackage{algorithm}
\usepackage{algorithmic}
\usepackage{booktabs}
\usepackage{amsmath}
\usepackage{amssymb}
\usepackage{multirow}
\usepackage[table]{xcolor}
\title{Confusion-Geometry Rebalancing for Long-Tailed Adversarial Training}
\author{%
  Mengnan Zhao\textsuperscript{1},
  Geyong Min\textsuperscript{3},
  Lihe Zhang\textsuperscript{2},
  Tianhang Zheng\textsuperscript{4},
  Jie Cui\textsuperscript{1}
}
\affiliations{%
  \textsuperscript{1}School of Computer Science, Anhui University, China\\
  \textsuperscript{2}School of Information and Communication Engineering, Dalian University of Technology, China\\
  \textsuperscript{3}Department of Computer Science, University of Exeter, UK\\
  \textsuperscript{4}College of Computer Science, Zhejiang University, China
}

\begin{document}

\maketitle

\begin{abstract}
Adversarial training under long-tailed distributions suffers from a dual imbalance: the class imbalance skews the training objective toward head classes, and the adversarial inner maximization may further amplify this bias. Existing methods mitigate this issue by correcting class priors or adapting class-wise robust supervision, yet they treat each class in isolation and fail to identify which boundaries drive long-tailed collapse. We propose a Confusion-Geometry Rebalancing method (CGRm) for long-tail  adversarial training, a plug-in framework that leverages directed robust errors as training signals. 
CGRm leverages periodic robust evaluations to derive source‑class loss weights, class‑wise robust coefficients, and a directed confusion‑geometry graph. The method then couples feedback‑weighted robust optimization with graph‑guided margin correction, thereby boosting the robustness of vulnerable classes and sharpening the critical boundaries that drive long‑tailed performance degradation.
Experiments on long-tailed benchmarks show that CGRm achieves consistent robust performance gains over existing methods, with ablations validating the contribution of each component. We provide the code in the supplement.
\end{abstract}

\begin{figure}[t]
    \centering
    \includegraphics[width=1\linewidth]{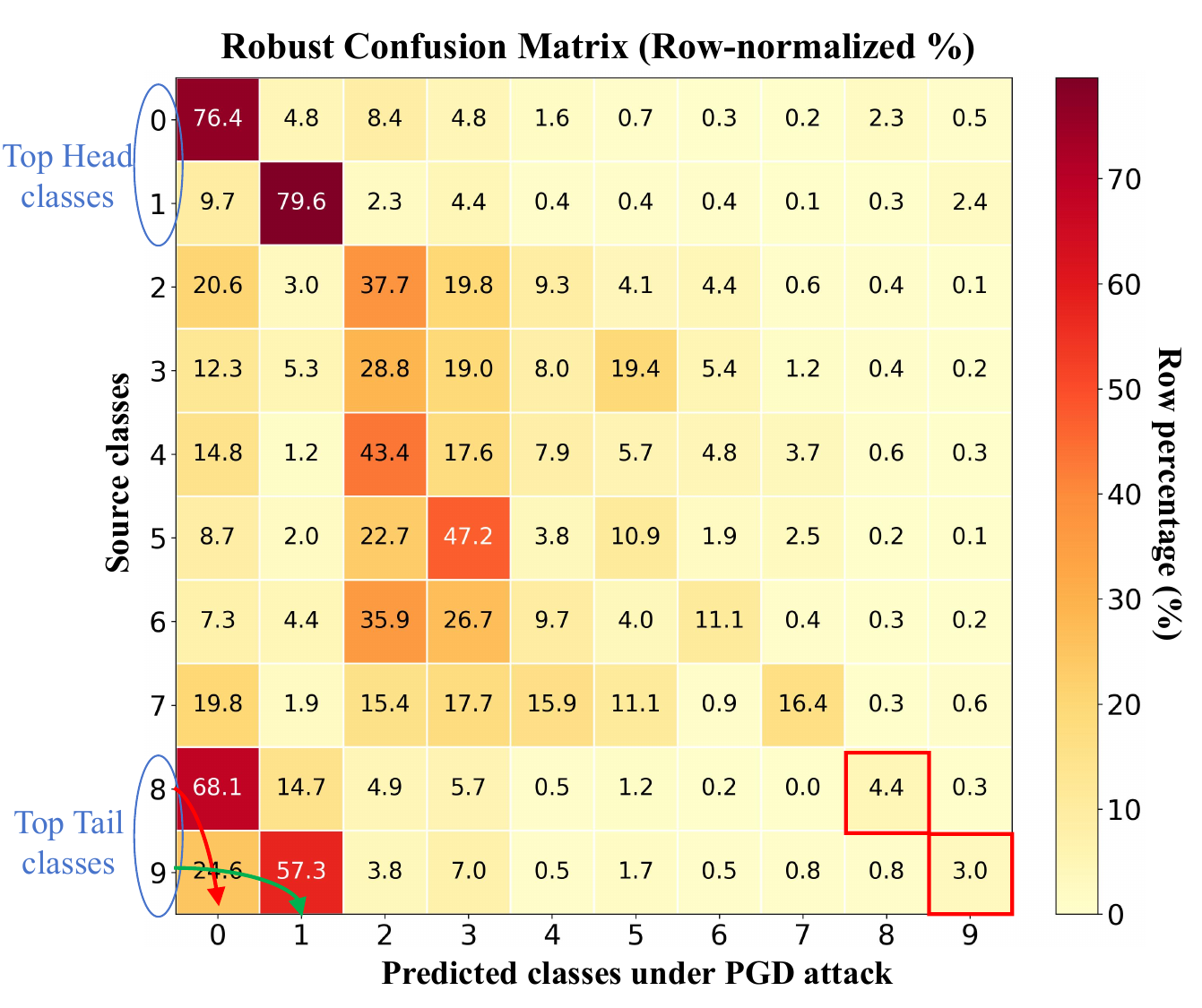}
    \caption{Robust confusion matrix on CIFAR10-LT. Adversarial examples from tail classes are frequently misclassified to head classes, often in a directed manner, highlighting the importance of targeted margin correction.}
    \label{fig1}
\end{figure}

\section{Introduction}
The vulnerability of deep neural networks to imperceptible adversarial perturbations has raised critical concerns about their deployment in safety-critical systems. Adversarial training (AT) is widely regarded as one of the most effective defenses against this threat. It is typically formulated as a min-max optimization problem \citep{zhao2024atsurvey,zhao2023smooth,zhao2024coblessing,zhao2026erroramplification}. Despite its empirical success, AT has been predominantly studied under an implicit assumption of balanced class priors. However, this assumption rarely holds in real-world scenarios, where sampled datasets are intrinsically long-tailed: a few head classes account for the majority of samples, while numerous tail classes are represented by very few instances \citep{du2024probabilistic}. When AT is applied directly to such skewed data, it introduces a two-fold imbalance: head classes not only dominate the training objective, but also disproportionately influence the adversarial decision boundaries during the inner maximization \cite{jiang2026rethinking,qin2026fedcart,li2025head}. Consequently, the learned model may achieve satisfactory average robust accuracy, yet its performance on tail classes remains notably inferior.

Recent studies improve the long-tailed AT from complementary perspectives. One line of work focuses on prior correction, adjusting logits, losses, or sample weights to counteract label imbalance \citep{ren2020balanced,wu2021robal,yue2024revisiting}. Another thread develops class-aware robust calibration, adapting perturbation budgets, regularization strengths, or class-specific weights based on each class's robust behavior \citep{wei2023cfa,lee2024dafa}. Most recently, RobustLT shows that the adversarial inner maximization itself should also be class-dependent, reallocating perturbation radii and attack schedules across the imbalance spectrum \citep{zhang2026robustlt}. Despite these advances, existing methods share a common limitation: each class's robust error is treated in isolation, without considering where those errors are directed, as shown in Figure \ref{fig1}.

This directional information matters in long-tailed AT. Under adversarial attack, an example from a source class is driven across the decision boundary toward different target classes. In long-tailed recognition, these errors are often structured rather than random: tail-class examples tend to collapse into frequent classes, especially when their feature representations lie close in the adversarial space. Consequently, treating all errors from an under-robust class equally may waste robust supervision on irrelevant negatives, while scalar class feedback alone cannot identify which target classes actually absorb these errors. We argue that robust long-tailed training should consider not only whether a class is vulnerable, but also where its adversarial examples are directed and whether the resulting confusion is geometrically plausible.

Motivated by this observation, we propose a Confusion-Geometry Rebalancing method (CGRm) for long-tail AT, a plug-in framework that explicitly accounts for directed robust errors. The method periodically evaluates the model under attack and collects three forms of feedback—class-wise robust accuracy, confusion counts, and adversarial feature centers—which jointly characterize each class's vulnerability and its error direction. This feedback serves two complementary purposes. On one hand, it produces class-adaptive weights and robust coefficients that amplify the contribution and regularization of under-robust classes. On the other hand, it builds a directed confusion-geometry graph whose edges capture tail-to-head misclassifications that are both frequent and geometrically plausible; a margin loss is then applied exclusively to these selected pairs. In this way, CGRm rebalances not only how much each class is supervised, but also which boundaries are explicitly hardened.

Our contributions are summarized as follows:
\begin{itemize}
    \item We formulate long-tailed AT as a directed robust error-correction problem and propose CGRm, modeling not only per-class vulnerability but also the target direction and geometric plausibility of robust confusions.
    \item We introduce a dynamic feedback mechanism that decouples per-class gradient contribution from clean-robust regularization by separately updating loss weights and robust coefficients based on evolving robust-error statistics.
    \item We design a confusion-geometry boundary correction module that builds a directed graph from confusion frequency, target-class headness, and adversarial feature affinity, and then applies a margin loss exclusively to systematic source-target confusions.
    \item Extensive experiments on long-tailed benchmarks show that CGRm consistently outperforms existing methods.
\end{itemize}

\section{Related Work}

\textbf{Adversarial training (AT).}
AT is typically formulated as a min-max optimization problem \cite{goodfellow2015explaining,robey2024adversarial,yue2023revisiting,jia2024improving}.
Let $f_\theta$ be a classifier, $\ell$ be the loss, and $\mathcal{S}_\epsilon(x)=\{x':\|x'-x\|_\infty\le\epsilon\}$ be the perturbation set around an input $x$. Standard AT minimizes the worst-case risk,
\begin{equation}
    \min_\theta\;
    \mathbb{E}_{(x,y)\sim \mathcal{D}_{\mathrm{train}}}
    \left[
    \max_{x'\in\mathcal{S}_\epsilon(x)}
    \ell(f_\theta(x'),y)
    \right],
\end{equation}
where the inner maximization is usually approximated by PGD \citep{madry2018towards}. TRADES \citep{zhang2019trades} decouples the learning objective into a standard classification term and a robust regularization term:
\begin{equation}
    \mathbb{E}\left[
    \ell(f_\theta(x),y)
    +
    \beta
    \max_{x'\in\mathcal{S}_\epsilon(x)}
    D_{\mathrm{KL}}
    (p_\theta(\cdot|x)\|p_\theta(\cdot|x'))
    \right],
\end{equation}
where $p_\theta(\cdot|x)$ denotes the predictive distribution and $\beta$ governs the trade-off between clean accuracy and adversarial robustness. 
AWP \citep{wu2020awp} improves robust generalization by perturbing both inputs and model weights during training, expressed as 
\begin{equation}
\min_\theta \max_{\|v\|\le\gamma\|\theta\|}\mathbb{E}_{(x,y)\sim D}\left[ \max_{x'\in\mathcal{S}_\epsilon(x)}\ell(f_{\theta+v}(x'), y)\right],
\end{equation}
where $v$ denotes the weight perturbation and $\gamma$ controls the perturbation magnitude. 

\textbf{Long-tailed adversarial training.}
Unlike standard AT, long-tailed AT inherits two coupled biases: 1) the empirical risk is dominated by head classes, and 2) the adversarial examples generated around the learned decision boundary tend to further amplify head-class preferences.

One family of methods addresses long-tailed learning as a prior-correction problem, adjusting losses, logits, or sample weights to reduce the dominance of class frequency in robust training \citep{cao2019ldam,ren2020balanced}.
Given an adversarial example \(x' \in \mathcal{S}_\epsilon(x)\) and logits \(z = f_\theta(x')\), RoBal \citep{wu2021robal} integrates AT with dynamic logit adjustment, 
\begin{equation}\label{eq4}
\mathcal{L}_{\mathrm{RoBal}} = -\log \frac{\exp(z_y + \Delta_y)}{\sum_j \exp(z_j + \Delta_j)},
\end{equation}
where \(\Delta_j\) is estimated from per-class robust confusion patterns during training.
Instead of relying on dynamic confusion statistics, BSL \cite{yue2024revisiting} uses static class frequencies as the correction prior, formulated as
\begin{equation}
\mathcal{L}_{\mathrm{BSL}} = -\log \frac{n_y^{\tau_b}\exp(z_y)}{\sum_i n_i^{\tau_b}\exp(z_i)},
\end{equation}
where \(\tau_b\) tunes the strength of class-prior correction.
TAET \citep{wang2025taet} argues that static frequency correction may still overfit underrepresented or hard classes. It therefore introduces a two-stage adversarial equalization strategy. Stage I trains the model with standard cross-entropy to stabilize clean representations. Stage II optimizes adversarial examples with a hierarchical equalization loss.
Similarly, REAT \citep{li2023reat} down-weights the loss contribution of adversarial examples that are frequently predicted as head classes, encouraging the adversary to focus more on underrepresented categories. 

Beyond prior-correction, another line of work adapts robustness mechanisms on a per-class basis, such as perturbation radii and regularization coefficients.
CFA \citep{wei2023cfa} scales perturbation radii and regularization coefficients according to each class's current robust accuracy \(r_c\):
\begin{equation}
\epsilon_c = (\lambda_1 + r_c)\epsilon, \qquad
\beta_c = \frac{(\lambda_2 + r_c)\beta}{1 + (\lambda_2 + r_c)\beta},
\end{equation}
where \(\epsilon\) and \(\beta\) are base values, and \(\lambda_1, \lambda_2\) are hyperparameters. This per-class scaling relies on a scalar accuracy summary per class but does not capture cross-class confusion. UDR \citep{bui2022udr} instead abstracts the issue to the distribution level, formulating AT as Wasserstein distributional robustness. 
RobustLT \citep{zhang2026robustlt} tackles long-tailed AT through a perturbation rebalancing mechanism coupled with an adversarial iteration weighting schedule. For each class \(c\), it defines the imbalance ratio \(K_c = n_{\max} / n_c\) and formulates the maximum perturbation budget as:
\begin{equation}\label{eq14}
\bar{\epsilon}_c =
\left(1 - \alpha + \frac{\alpha \sqrt{\log K_c}}{\sum_j \frac{n_j}{N} \sqrt{\log K_j}}\right) \epsilon,
\end{equation}
where \(n_c\) denotes the sample count of class \(c\), $n_{\max}=\max_c n_c$, and the hyperparameter \(\alpha\) governs the extent to which perturbation budgets are redistributed from head to tail classes. The method further introduces a progressive warm-up schedule to enable more effective application of these budgets. At epoch \(t\), the actual perturbation radius and the corresponding PGD step size are given by
\begin{equation}\label{eq15}
\epsilon_c^{(t)} = \min\left(\frac{t-1}{\rho T}, 1\right)\bar{\epsilon}_c, \quad
\eta_c^{(t)} = \frac{\epsilon_c^{(t)}}{\epsilon}\eta,
\end{equation}
with \(T\) denoting the total number of epochs, \(\eta\) the base step size, and \(\rho\) the warm-up ratio.

\textbf{Difference from existing methods.}
Existing methods treat all misclassifications from a vulnerable class equally, ignoring both which competing class the error goes to and whether that confusion is geometrically plausible in feature space.
CGRm addresses these two missing dimensions. It jointly tracks robust errors and their source-target geometry: class frequency only initializes adversarial exploration, while subsequent corrections are driven by observed robust errors, their target classes, and the feature-space proximity of the confused classes. 

\section{Method}


This work propose CRGm, which periodically evaluates the model under attack and translates the robust errors into optimization objectives: class-adaptive weights and coefficients that rebalance per-class loss contribution and regularization toward clean-robust consistency, and a directed graph that applies margin loss to systematic source-target confusions. 

\subsection{Optimization with Error Feedback}
To mitigate the adverse effects of uniformly assigned perturbation budgets in imbalanced data scenarios, we reuse Eqs. (\ref{eq14}) and (\ref{eq15}) to calculate the attack budgets $\epsilon_c^{(t)}$ and step size $\eta_c^{(t)}$.
For a sample $(x,y)$, we then generate $x^{adv}$ using such class-conditioned budget $\epsilon_y^{(t)}$ and step size $\eta_y^{(t)}$. 

Given these adversarial examples, CGRm optimizes three coupled signals: a prior-calibrated natural classification term for long-tailed recognition, a feedback-weighted robust consistency term for vulnerable source classes, and a confusion-geometry margin term for the target classes that absorb adversarial errors. The objective is expressed as
\begin{equation}
    \mathcal{L}_{\mathrm{CGR}}
    =
    \mathbb{E}\left[\ell_{\mathrm{bal}}(f_\theta(x),y)\right]
    +
    \mathcal{L}_{\mathrm{rob}}
    +
    \lambda_m \mathcal{L}_{\mathrm{cgm}} .
\label{eq:objective}
\end{equation}
The first term denotes a balanced prior-calibrated classification term. Let $\pi_c=n_c/N$ and $b_c=\log(\pi_c)$. For logits $z=f_\theta(x)$, we use
\begin{equation}
    \ell_{\mathrm{bal}}(z,y)
    =
    -\log
    \frac{\exp(z_y + b_y)}
    {\sum_{j=1}^{C}\exp(z_j+ b_j)}.
\label{eq:balanced-ce}
\end{equation}
$\mathcal{L}_{\mathrm{rob}}$ reallocates the consistency between clean and adversarial predictions across classes. We introduce two independent factors: the source-class weight \(w_y^{(t)}\) modulates the overall gradient contribution from samples of class \(y\), while the class-wise robust coefficient \(\beta_y^{(t)}\) controls the strength of the consistency constraint imposed on that class. This allows the model to emphasize an under-robust class and independently enforce a stronger clean-adversarial prediction agreement on it. For example, when adopting TRADES as the base learner, $\mathcal{L}_{\mathrm{rob}}$ is expressed as
\begin{equation}\label{eq18}
    \mathcal{L}_{\mathrm{rob}}
    =
    \mathbb{E}
    \left[
    w_y^{(t)}\beta_y^{(t)}
    D_{\mathrm{KL}}
    \left(
    f_\theta(x)\;\|\;f_\theta(x^{adv})
    \right)
    \right].
\end{equation}
where $D_{\mathrm{KL}}$ penalizes the discrepancy between clean and adversarial predictions. 
The confusion-geometry margin complements this source-class correction with target-class information. A low robust accuracy indicates that class $y$ is vulnerable, but it does not identify which wrong class should be separated from $y$. The graph $G^{(t)}$ selects target classes that adversarial examples from $y$ are frequently confused with and that are geometrically plausible competitors. CGRm then applies a margin penalty only to these selected targets:
\begin{equation}
    \mathcal{L}_{\mathrm{cgm}}
    =
    \mathbb{E}_{(x,y)}
    \left[
    w_y^{(t)}
    \sum_{j\ne y}
    G_{yj}^{(t)}
    \left[m-z_y^{adv}+z_j^{adv}\right]_+
    \right],
\label{eq:graph-margin}
\end{equation}
where $m$ denotes the margin threshold, $z^{adv}=f_\theta(x^{adv})$. At the beginning of training, $w_y=1$, $\beta_y=\beta$, and $G_{y,j}=0$. As training progresses, these variables are updated from robust-error statistics, which we define next.

\subsection{Robust-Error Statistics}

The frequency prior is only a proxy for robust difficulty. Some rare classes may be clearly separable, whereas some medium-frequency classes may remain vulnerable due to feature overlap or semantic ambiguity. CGRm thus updates its training emphasis using the model's own robust behavior. At an evaluation epoch $t$, we attack a monitoring split (subset of the training dataset) and compute the class-wise robust accuracy $a_c^{(t)}$ and robust confusion matrix $M^{(t)}$, where $M_{ij}^{(t)}$ counts examples from class $i$ predicted as class $j$ under attack.

We convert these measurements into a feedback score using two complementary signals. The first measures how far a class falls below the mean robust accuracy,
\begin{equation}
    g_c^{(t)} = \max\left(\bar{a}^{(t)}-a_c^{(t)},0\right),
    \qquad
    \bar{a}^{(t)}=\frac{1}{C}\sum_{j=1}^{C}a_j^{(t)}.
\label{eq:robust-gap}
\end{equation}
The second measures how often adversarial examples from class $c$ leave the correct class:
\begin{equation}
    e_c^{(t)} =
    1 -
    \frac{M_{cc}^{(t)}}{\sum_{j=1}^{C} M_{cj}^{(t)}+\xi},
\label{eq:conf-error}
\end{equation}
where \(\xi\) is a negligible positive constant for numerical stability, \(M_{ij}^{(t)}\) counts examples from source class \(i\) that are predicted as target class \(j\) under attack.
We combine the two signals as $s_c^{(t)}$,
\begin{equation}
    s_c^{(t)} =
    \frac{g_c^{(t)}+ e_c^{(t)}}
    {\frac{1}{C}\sum_{j=1}^{C}(g_j^{(t)}+ e_j^{(t)})+\xi}.
\label{eq:feedback-score}
\end{equation}
Because robust evaluation can be noisy, especially for tail classes, we use an exponential moving average
\begin{equation}
    \tilde{s}_c^{(t)}
    =
    \mu \tilde{s}_c^{(t-1)} + (1-\mu)s_c^{(t)}.
\label{eq:feedback-ema}
\end{equation}
Based on $\tilde{s}_c^{(t)}$, we calculate the per-class loss weight in Eq. (\ref{eq18}), which answers the question of how much examples from class $c$ should contribute to the optimization objective:
\begin{equation}
    w_c^{(t)} =
    \mathrm{clip}
    \left(
    \frac{1+\tilde{s}_c^{(t)}}
    {\frac{1}{C}\sum_{j=1}^{C}(1+ \tilde{s}_j^{(t)})},
    0,w_{\max}
    \right).
\label{eq:class-weight}
\end{equation}
Clipping prevents noisy tail estimates from producing excessive gradients, while normalization keeps the average weight stable across epochs.

Additionally, \(\tilde{s}_c^{(t)}\) is used to determine the regularization strength for each class. A single global coefficient \(\beta\) applies the same clean-robust trade-off across all classes, which becomes suboptimal when robust difficulty varies unevenly: increasing \(\beta\) universally may hurt clean accuracy on easier classes, while decreasing it leaves harder classes under-regularized. To address this, we adapt \(\beta\) per class based on the proposed robust-error score:
\begin{equation}
\beta_c^{(t)} = \beta \cdot \frac{ \mathrm{clip}\left(1+\tilde{s}_c^{(t)}/\bar{s}^{(t)}, 1 ,r_{\max}\right)} {\frac{1}{C}\sum_{j=1}^{C} \mathrm{clip}\left(1+ \tilde{s}_j^{(t)}/\bar{s}^{(t)}, 1,r_{\max}\right)},
\label{eq:class-beta}
\end{equation}
where \(\bar{s}^{(t)}=\frac{1}{C}\sum_j \tilde{s}_j^{(t)}\). Unlike \(w_c^{(t)}\), which modulates a class's contribution to the overall objective, \(\beta_c^{(t)}\) adjusts the clean-robust trade-off within the robust term. 
\begin{algorithm}[t]
\caption{CGRm Training}
\label{alg:cgrlt}
\begin{algorithmic}[1]
\STATE \textbf{Input:} long-tailed training set $\mathcal{D}$, class counts $\{n_c\}_{c=1}^{C}$, model $f_\theta$, base attack parameters $(\epsilon,\eta)$, base robust coefficient $\beta$, total epochs $T$.
\STATE \textbf{Output:} robust model $f_\theta$.
\STATE Initialize feedback scores $\tilde{s}_c^{(0)}=1$, class weights $w_c^{(0)}=1$, robust coefficients $\beta_c^{(0)}=\beta$, and graph $G^{(0)}=0$;
\FOR{epoch $t=1,\ldots,T$}
    \STATE Obtain class-wise perturbation budgets $\epsilon_c^{(t)}$ and step sizes $\eta_c^{(t)}$ from Eq.~\eqref{eq15};
    \FOR{mini-batch $(x,y)$ sampled from $\mathcal{D}$}
        \STATE Generate $x^{adv}$ using $\epsilon_y^{(t)}$ and $\eta_y^{(t)}$;
        \STATE Compute prior-calibrated loss $\ell_{\mathrm{bal}}$ by Eq.~\eqref{eq:balanced-ce};
        \STATE Compute the feedback-weighted robust loss $\mathcal{L}_{\mathrm{rob}}$ with $w_y^{(t-1)}$ and $\beta_y^{(t-1)}$ by Eq.~\eqref{eq18};
        \STATE Compute the confusion-geometry margin $\mathcal{L}_{\mathrm{cgm}}$ with $G^{(t-1)}$ by Eq.~\eqref{eq:graph-margin};
        \STATE Update $\theta$ by minimizing $\mathcal{L}_{\mathrm{CGR}}$ in Eq.~\eqref{eq:objective};
    \ENDFOR
    \IF{$t$ is an evaluation epoch}
        \STATE Attack a monitoring split to compute robust gaps $g_c^{(t)}$, confusion errors $e_c^{(t)}$, and smoothed feedback scores $\tilde{s}_c^{(t)}$ by Eqs.~\eqref{eq:robust-gap}--\eqref{eq:feedback-ema};
        \STATE Update $w_c^{(t)}$ and $\beta_c^{(t)}$ based on $e_c^{(t)}$ and $\tilde{s}_c^{(t)}$;
        \STATE Compute robust confusion $P^{(t)}$, headness $h$, and feature affinity $A^{(t)}$ by Eqs.~\eqref{eq27}--\eqref{eq29};
        \STATE Update the smoothed confusion-geometry graph $G^{(t)}$ by Eq.~\eqref{eq30};
    \ENDIF
\ENDFOR

\end{algorithmic}
\end{algorithm}

\renewcommand{\arraystretch}{0.93}
\begin{table*}[htpb]
\centering
\caption{Natural and robust accuracies of various base AT algorithms using ResNet. Bold digits denotes the best result.}
\label{tab1}
\resizebox{\textwidth}{!}{
\begin{tabular}{lcccccccccccc}
\toprule
\multirow{2}{*}{\bf Methods}
& \multicolumn{4}{c}{\bf CIFAR10-LT}
& \multicolumn{4}{c}{\bf CIFAR100-LT}
& \multicolumn{4}{c}{\bf TinyImageNet-LT} \\
\cmidrule(lr){2-5}\cmidrule(lr){6-9}\cmidrule(lr){10-13}
& Nat.(all) & Nat.(tail) & Rob.(all) & Rob.(tail)
& Nat.(all) & Nat.(tail) & Rob.(all) & Rob.(tail)
& Nat.(all) & Nat.(tail) & Rob.(all) & Rob.(tail) \\
\midrule

\rowcolor{gray!15} {REAT} &69.59\% & 63.79\% & 31.41\% & 21.91\% &45.49\% & 43.35\% & 17.54\% & 16.59\% &37.91\% & 35.42\% & 12.80\% & 11.33\%\\
{TAET} & 65.45\% & 59.80\% & 33.32\% & 26.12\% & 43.65\% & 40.05\% & 17.36\% & 15.94\% & 34.99\% & 31.77\% & 12.04\% & 10.29\%\\
 \rowcolor{gray!15}{RoBal}& 72.43\% & 68.14\% & 33.47\% & 26.71\% & 44.88\% & 43.79\% & 18.61\% & 18.10\% &37.12\% & 35.95\% & 12.28\% & 11.56\% \\
{BSL } &70.04\% & 64.35\% & 31.98\% & 22.75\% & 45.90\% & 43.53\% & 18.10\% & 17.10\% &38.90\% & 36.36\% & 13.08\% & 11.41\%\\

 \midrule
 \rowcolor{gray!15}PGD-AT & 56.33\% & 46.27\% & 26.28\% & 13.19\% & 43.03\% & 38.91\% & 15.40\% & 13.34\% & 34.80\% & 30.56\% & 9.16\% & 7.25\% \\
 + UDR &  57.60\% & 47.71\% & 29.71\% & 16.94\% & 42.58\% & 38.26\% & 15.72\% & 13.86\% &34.90\% & 29.86\% & 12.17\% & 9.64\%\\
 \rowcolor{gray!15}+ CFA & 57.08\% & 47.23\% & 29.76\% & 16.31\% & 43.31\% & 38.60\% & 17.58\% & 15.54\% & 37.08\% & 32.00\% & 11.96\% & 9.55\% \\
 + DAFA & 64.14\% & 56.91\% & 30.51\% & 21.26\% & 44.17\% & 40.39\% & 16.78\% & 15.55\%&36.39\% & 32.19\% & 12.22\% & 10.35\% \\
 \rowcolor{gray!15}+ RobustLT & 56.71\% & 46.81\% & 30.88\% & 18.60\% & 46.09\%  & 42.14\% & 17.73\% & 16.01\%&36.67\% & 31.89\% & 12.26\% & 10.08\% \\
 + Ours & \textbf{71.80\%} & \textbf{68.15\%} & \textbf{35.31\%} & \textbf{27.93\%} & \textbf{48.25\%} & \textbf{46.66\%} & \textbf{18.83\%} & \textbf{18.19\%} & \textbf{39.81\%} & \textbf{38.01\%} & \textbf{13.03\%} & \textbf{11.83\%}\\
\midrule
 \rowcolor{gray!15}TRADES & 54.48\%  & 43.91\% & 30.13\% & 16.66\% &  43.57\% & 38.51\%& 20.07\% &17.25\%  &  35.41\% & 30.18\% & 13.85\% & 10.78\%  \\
 + UDR & 55.09\% &44.64\%&29.79\% & 16.59\%  & 41.01\%  &36.08\% & 19.21\% & 16.93\% & 33.72\% & 28.73\% & 12.61\% & 9.85\%  \\
 \rowcolor{gray!15}+ CFA & 53.32\%  & 42.43\%& 30.45\% & 16.41\% & 43.07\% & 37.52\% & 21.16\% & 18.18\% & 37.60\% & 32.48\% & 14.82\% & 11.61\%   \\
 + DAFA & 61.25\% & 54.24\% & 31.54\% & 21.93\% &44.77\% & 40.83\% & 19.43\% & 18.04\% &36.07\% & 32.21\% & 13.21\% & 11.19\% \\
 \rowcolor{gray!15}+ RobustLT & 56.44\% & 44.01\% & 31.38\% & 18.57\% & 45.50\% & 40.80\% & 20.68\% &18.25\% &36.46\% & 31.66\% & 13.78\% &10.82\% \\
 + Ours & \textbf{68.29\%} & \textbf{62.39\%} & \textbf{34.92\%} & \textbf{27.06\%} & \textbf{48.81\%} & \textbf{46.80\%} & \textbf{21.54\%} & \textbf{20.73\%} & \textbf{38.84\%} & \textbf{36.82\%} & \textbf{14.12\%}& \textbf{13.05\%} \\
\midrule
\rowcolor{gray!15}AWP &48.04\% & 35.68\% & 31.50\% & 17.77\% &43.45\% & 37.76\% & 22.72\% & 19.45\%&34.27\% & 28.32\% & 16.73\% & 12.95\%   \\
 + UDR &43.59\% & 30.28\% & 28.00\% &14.22\% &43.58\% & 38.45\% & 21.08\% & 18.25\% &31.54\% & 25.79\% & 14.10\% & 10.78\% \\
 \rowcolor{gray!15}+ CFA &53.28\% & 42.38\% & 32.51\% &18.81\% &46.72\% & 41.71\% & 21.81\% & 18.93\% &38.56\% & 32.94\% & 16.02\% & 12.53\%\\
 + DAFA &52.67\% & 42.66\% & 32.45\% & 22.45\% &44.50\% & 39.30\% & 23.03\% & 20.42\% &35.03\% & 29.78\% & 16.99\% & 13.94\% \\
 \rowcolor{gray!15}+ RobustLT &57.74\% & 44.35\% & 36.69\% & 27.06\% &45.67\% & 39.65\% & 22.96\% & 19.86\%& 34.38\%& 28.10\% & 16.50\% & 12.62\%
\\
 + Ours &\textbf{68.19\%} & \textbf{64.71\%} & \textbf{40.70\%} & \textbf{36.44\%} & \textbf{50.30\%} & \textbf{48.67\%} & \textbf{24.22\%} & \textbf{23.85\%} & \textbf{39.99\%} & \textbf{37.65\%} & \textbf{18.43\%} & \textbf{17.09\%}\\
\bottomrule
\end{tabular}}
\end{table*}

\subsection{Confusion-Geometry Boundary Correction}
Class-wise feedback identifies vulnerable source classes, yet it treats each class independently, overlooking the inherently relational nature of robust failure. An adversarial example from class \(i\) becomes harmful precisely because it crosses the decision boundary toward a specific target class \(j\). In long-tailed recognition, the most detrimental errors are often not random misclassifications, but rather tail classes being absorbed by neighboring head classes. To address this, CGRm constructs a directed confusion-geometry graph that selects source-target pairs requiring explicit margin correction.

The graph integrates three complementary signals: empirical confusion probabilities, class-frequency bias, and feature-geometry affinity. The graph at epoch \(t\) is formulated as the element-wise product of these three terms:
\begin{equation}\label{eq26}
\hat{G}_{ij}^{(t)} = P_{ij}^{(t)} \, h_j \, A_{ij}^{(t)}, \qquad i \ne j, \qquad \hat{G}_{ii}^{(t)} = 0.
\end{equation}
Here, the row-normalized robust confusion probability \(P_{ij}^{(t)}\) reflects the empirical tendency of adversarial examples from \(i\) to cross the decision boundary toward \(j\).
\begin{equation}\label{eq27}
P_{ij}^{(t)} = \frac{M_{ij}^{(t)}}{\sum_{k=1}^{C}M_{ik}^{(t)}+\xi}.
\end{equation} 
The headness score \(h_j\) aims to penalize confusions that are more damaging under long-tailed imbalance:
\begin{equation}\label{eq28}
h_j = \sqrt{\frac{n_j}{n_{\max}}},
\end{equation}
where \(n_c\) is the sample count of class \(c\) and \(n_{\max} = \max_c n_c\). 
The geometry affinity \(A_{ij}^{(t)}\) measures the geometric proximity between the two classes in the adversarial feature space. Let $m_c^{(t)}$ denote the feature center of class $c$ computed from adversarial examples, and define 
\begin{equation}\label{eq29}
    A_{ij}^{(t)}
    =
    \frac{1}{\| \hat{m}_i^{(t)}-\hat{m}_j^{(t)}\|_2+\xi},
\end{equation}
where $\hat{m}$ is $\ell_2$-normalized.

To reduce noise and maintain sparsity, we retain only the top-\(k\) outgoing edges for each source class and normalize the nonzero entries to unit mean. The graph is then updated smoothly across epochs via exponential moving average:
\begin{equation}\label{eq30}
G^{(t)} = \mu G^{(t-1)} + (1-\mu)\hat{G}^{(t)}.
\end{equation}
The target classes selected by Eq.~\eqref{eq26} are those that (i) are frequently confused with adversarial examples from the source class, as measured by Eqs.~\eqref{eq27} and \eqref{eq28}, and (ii) serve as geometrically plausible competitors, as quantified by Eq.~\eqref{eq29}. Thus, this graph serves as a targeted description of long-tailed robust collapse.

The training procedure is summarized in Algorithm~\ref{alg:cgrlt}.

\section{Experiments}

\renewcommand{\arraystretch}{0.93}
\begin{table*}[t]
\centering
\caption{Natural and robust accuracies of various AT algorithms using PreActResNet.}
\label{tab2}
\resizebox{\textwidth}{!}{
\begin{tabular}{lcccccccccccc}
\toprule
\multirow{2}{*}{\bf Methods}
& \multicolumn{4}{c}{\bf CIFAR10-LT}
& \multicolumn{4}{c}{\bf CIFAR100-LT}
& \multicolumn{4}{c}{\bf TinyImageNet-LT} \\
\cmidrule(lr){2-5}\cmidrule(lr){6-9}\cmidrule(lr){10-13}
& Nat.(all) & Nat.(tail) & Rob.(all) & Rob.(tail)
& Nat.(all) & Nat.(tail) & Rob.(all) & Rob.(tail)
& Nat.(all) & Nat.(tail) & Rob.(all) & Rob.(tail) \\
\midrule
\rowcolor{gray!15}{REAT}  & 70.89\% & 65.21\% & 32.26\% & 23.04\% &45.46\% & 42.99\% & 17.88\% & 16.89\% & 38.40\% & 35.76\% & 12.94\% & 11.55\% \\
{TAET}  & 66.54\% & 61.24\% & 33.73\% & 26.88\% & 43.32\% & 39.42\% & 18.04\% & 16.35\% & 35.12\% & 31.70\% & 12.19\% & 10.44\%\\
\rowcolor{gray!15}{RoBal} & 72.27\% & 68.10\% & 34.45\% & 27.38\% &45.04\% & 44.38\% & 18.29\% & 17.61\% & 36.39\% & 35.25\% & 13.31\% & 12.26\%\\
{BSL } & 70.49\% & 64.88\% & 32.15\% & 22.85\% &45.13\% & 42.89\% & 17.60\% & 16.75\% & 38.31\% & 35.96\% & 13.31\% & 12.00\% \\
 \midrule
\rowcolor{gray!15}PGD-AT & 57.38\% & 47.64\% & 26.93\% & 13.85\% &42.67\% & 38.46\% & 15.36\% & 13.53\% & 34.39\% & 30.03\% & 9.34\% & 7.46\% \\
 + UDR & 59.62\% & 50.34\% & 29.69\% & 16.99\%&42.24\% & 37.85\% & 15.76\% & 13.95\% & 34.87\% & 30.20\% &10.59\% & 8.56\%  \\
 \rowcolor{gray!15}+ CFA &  57.30\% & 47.54\% & 30.28\% & 16.90\% &43.25\% & 38.75\% & 17.13\% & 15.06\% & 36.93\% & 32.06\% & 12.19\% & 9.71\% \\
 + DAFA &  66.05\% & 59.31\% & 31.36\% & 22.30\%&43.69\% & 40.08\% & 16.91\% & 15.72\%& 
37.13\% & 32.94\% & 12.49\% & 10.81\% \\
 \rowcolor{gray!15}+ RobustLT & 57.36\% & 47.59\% & 31.70\% & 19.46\% &46.24\% & 42.23\% & 17.88\% & 16.20\%&37.47\% & 32.86\% & 12.48\% & 10.29\%\\
 + Ours &\textbf{73.03\%} & \textbf{68.59\%} & \textbf{35.08\%} & \textbf{28.11\%} &\textbf{47.75\%} & \textbf{46.35\%} & \textbf{18.39\%} & \textbf{17.76\%} & 
\textbf{39.68\%} & \textbf{37.75\%} & \textbf{12.68\%} & \textbf{11.71\%}\\
\midrule
\rowcolor{gray!15}TRADES & 56.19\% & 46.09\% & 30.68\% & 17.32\%  &43.49\% & 38.52\% & 19.85\% & 17.29\% & 35.74\% & 30.56\% & 14.00\% & 10.96\% \\
 + UDR & 56.44\% & 46.38\% & 30.38\% & 17.15\%&40.78\% & 35.95\% & 18.82\% & 16.48\% &33.82\% & 28.50\% & 12.09\% & 9.21\%  \\
 \rowcolor{gray!15}+ CFA &54.39\% & 43.70\% & 30.51\% &16.59\%&43.53\% & 38.01\% & 20.66\% & 17.65\% &
37.32\% & 31.79\% & \textbf{14.60\%} & 11.09\%\\
 + DAFA &65.06\% & 58.35\% &32.13\% & 22.51\%&44.62\% & 41.09\% & 19.69\% & 18.31\% &36.61\% & 32.59\% & 13.46\% & 11.43\%\\
 \rowcolor{gray!15}+ RobustLT & 57.33\% & 47.55\% & 33.12\% &22.57\%&45.21\% & 40.33\% & 21.22\% & 18.88\%&36.25\% & 31.08\% & 14.08\% & 11.22\% \\
 + Ours & \textbf{69.15\%} & \textbf{62.59\%} & \textbf{35.28\%} & \textbf{26.03\%} & \textbf{49.09\%} & \textbf{46.67\%} & \textbf{21.23\%} & \textbf{20.21\%} & \textbf{39.78\%} & \textbf{38.08\%} & 13.96\% & \textbf{13.28\%}\\
\midrule
 \rowcolor{gray!15}AWP &50.78\% & 39.05\% & 32.18\% & 18.50\% &43.53\% & 37.86\% & 23.04\% & 19.74\% & 33.99\% & 28.20\% & 16.96\% & 13.31\%\\
 + UDR & 46.16\% & 33.46\% & 29.64\% & 15.81\% &45.60\% & 40.65\% & 21.17\% & 18.60\% &34.39\% & 28.78\% & 15.44\% & 12.10\%\\
 \rowcolor{gray!15}+ CFA &54.86\% & 44.26\% & 32.84\% & 19.21\% &47.06\% & 41.93\% & 22.19\% & 19.25\% &38.63\% & 32.85\% & 16.35\% & 12.90\%\\
 + DAFA &56.78\% & 47.60\% & 33.48\% & 23.60\% &44.62\% & 39.59\% & 22.89\% & 20.38\% &35.15\% & 29.81\% & 16.74\% & 13.75\% \\
 \rowcolor{gray!15}+ RobustLT &45.82\% & 32.91\% & 31.52\% & 18.05\%&45.87\% & 39.95\% & 23.20\% & 20.29\%&34.65\% & 28.16\% & 16.70\% & 12.97\% \\
 + Ours & \textbf{69.59\%} & \textbf{66.25\%} & \textbf{41.42\%} & \textbf{37.25\%} & \textbf{51.11\%} & \textbf{49.64\%} & \textbf{24.28\%} & \textbf{23.97\%}&\textbf{39.85\%} & \textbf{37.56\%} & \textbf{18.85\%} & \textbf{17.68\%}\\
\bottomrule
\end{tabular}}
\end{table*}

\renewcommand{\arraystretch}{0.935}
\begin{table*}[t]
\centering
\caption{Natural and robust accuracies of various base AT algorithms using WideResNet.}
\label{tab3}
\resizebox{\textwidth}{!}{
\begin{tabular}{lcccccccccccc}
\toprule
\multirow{2}{*}{\bf Methods}
& \multicolumn{4}{c}{\bf CIFAR10-LT}
& \multicolumn{4}{c}{\bf CIFAR100-LT}
& \multicolumn{4}{c}{\bf TinyImageNet-LT} \\
\cmidrule(lr){2-5}\cmidrule(lr){6-9}\cmidrule(lr){10-13}
& Nat.(all) & Nat.(tail) & Rob.(all) & Rob.(tail)
& Nat.(all) & Nat.(tail) & Rob.(all) & Rob.(tail)
& Nat.(all) & Nat.(tail) & Rob.(all) & Rob.(tail) \\
\midrule

\rowcolor{gray!15}{REAT} &67.37\% & 60.31\% & 30.19\% & 18.73\% &
48.08\% & 44.93\% & 19.48\% & 18.31\% & 42.30\% & 39.96\% & 15.19\% & 13.51\%\\
{TAET} &66.89\% & 60.40\% & 34.51\% & 25.25\% &45.58\% & 40.79\% & 19.07\% & 17.12\% & 38.26\% & 33.85\%&13.16\% & 10.87\%
\\
\rowcolor{gray!15}{RoBal} & 72.50\% & 67.71\% & 34.62\% & 25.77\%
&48.84\% & 47.38\% & 21.85\% & 21.20\% & 41.26\% & 40.05\% & 16.05\% & 15.06\%
\\
{BSL } &67.65\% & 60.61\% & 30.64\% & 19.09\% &47.92\% & 45.05\% & 20.11\% & 18.84\% &41.65\% & 36.90\% & 14.71\% & 12.17\% \\
\midrule
\rowcolor{gray!15}PGD-AT & 58.85\% & 49.33\% & 26.86\% & 13.12\% &46.20\% & 41.65\% & 17.22\% & 15.01\% & 37.74\% & 33.50\% & 10.53\% & 8.43\%\\
 + UDR & 59.44\% & 50.06\% & 28.11\% & 14.96\% & 45.47\% & 40.90\% & 17.29\% & 15.20\% & 39.23\% & 34.45\% & 12.83\% & 10.34\% \\
 \rowcolor{gray!15}+ CFA &57.52\% & 47.59\% & 30.07\% & 16.41\% & 45.44\% & 40.55\% & 19.19\% & 16.90\% & 38.69\% & 34.07\% & 10.94\% & 9.12\%\\
 + DAFA &65.06\% & 57.39\% & 29.60\% & 18.71\% & 46.87\%  & 43.23\% & 18.06\% & 16.89\% &39.78\% & 35.89\% & 13.80\% & 11.81\% \\
 \rowcolor{gray!15}+ RobustLT & 60.74\% & 51.60\% & 31.54\% & 19.11\%& 49.07\%  & 44.55\% & 18.55\% & 16.74\% &41.65\% & 36.90\% & 14.71\% & 12.17\%  
\\
 + Ours &\textbf{73.83\%} & \textbf{68.59\%} & \textbf{35.64\%} & \textbf{26.41\%} &\textbf{50.95\%}  & \textbf{48.85\%} & \textbf{19.54\%} & \textbf{18.78\%} & \textbf{44.14\%} & \textbf{42.45\%} & \textbf{14.77\%} & \textbf{13.74\%}
 \\
\midrule
\rowcolor{gray!15}TRADES& 59.04\%  & 49.53\% & 32.42\% & 19.09\%&47.60\% & 42.19\% & 22.10\% & 19.60\%&38.83\% & 33.60\% & 16.07\% & 12.72\%\\
 + UDR &58.73\%  & 49.09\% & 31.36\% & 18.07\%&44.95\% & 39.85\% & 20.32\% & 17.91\% & 40.85\% & 35.90\% & 13.67\% & 11.19\%\\
 \rowcolor{gray!15}+ CFA &55.00\% & 44.41\% & 31.85\% & 17.99\%&47.15\% & 41.98\% & 22.58\% & 19.62\%& 39.57\% & 34.14\% & 16.80\% & 13.11\%\\
 + DAFA & 65.58\% & 58.48\% & 33.54\% & 22.94\%&48.52\% & 44.84\% & 21.98\% & 20.45\% & 39.15\% & 34.91\% & 15.91\% & 13.45\%  \\
 \rowcolor{gray!15}+ RobustLT & 60.53\% & 51.36\% & 33.60\% & 21.00\% & 49.98\% & 45.32\% & 23.22\% & 21.12\% & 39.76\% & 34.76\% & 16.78\% & 13.71\%\\
 + Ours & \textbf{68.37\%} & \textbf{61.54\%} & \textbf{35.81\%} & \textbf{25.16\%} & \textbf{52.71\%} & \textbf{49.20\%} & \textbf{23.60\%} & \textbf{22.04\%}& \textbf{42.84\%} & \textbf{41.24\%} & \textbf{17.07\%} & \textbf{16.15\%} \\
\midrule
\rowcolor{gray!15} AWP & 53.39\% & 42.35\% & 31.21\% & 16.96\% & 50.03\% & 44.60\% & 24.81\% & 21.29\% & 39.10\% & 32.86\% & 14.87\% & 11.26\%\\
 + UDR & 57.87\% & 47.89\% & 32.68\% & 19.50\% &50.71\% & 45.74\% & 22.94\% & 20.26\% & {41.25\%} & 35.71\% & 12.68\%&10.74\%\\
\rowcolor{gray!15} + CFA&57.81\% & 47.77\% & 34.47\% & 20.70\% & 50.50\% & 45.25\% & 24.31\% & 21.30\% & 39.96\% & 33.48\% & 15.80\%&11.65\%\\
 + DAFA & 63.27\% & 55.31\% & 35.93\% & 25.94\% &49.78\% & 45.05\% & 25.09\% & 22.78\% & 35.13\% & 29.71\% & 16.95\% & 13.96\%\\
\rowcolor{gray!15} + RobustLT &68.91\% & 56.60\% & 38.70\% & 27.07\% & 50.83\% & 45.16\% & 25.34\% & 22.64\% & 38.01\% & 31.29\% & 19.29\% & 15.31\%\\
 + Ours & \textbf{74.07\%} & \textbf{69.98\%} & \textbf{43.72\%} & \textbf{37.79\%} & \textbf{55.95\%} & \textbf{54.60\%} & \textbf{26.06\%} & \textbf{25.85\%} 
 & \textbf{43.99\%}& \textbf{41.79\%} & \textbf{21.40\%}& 
 \textbf{20.13\%}\\
\bottomrule
\end{tabular}}
\end{table*}

\renewcommand{\arraystretch}{0.94}
\begin{table*}[h]
\centering
\caption{Robust accuracies against different attacks using WRN-28-10. Better results are bolded.}
\label{tab:attack-wrn}
\resizebox{\textwidth}{!}{
\begin{tabular}{llcccccccccccc}
\toprule
\multirow{2}{*}{\bf Methods}
& \multicolumn{4}{c}{\bf CIFAR10-LT}
& \multicolumn{4}{c}{\bf CIFAR100-LT}
& \multicolumn{4}{c}{\bf TinyImageNet-LT} \\
\cmidrule(lr){2-5}\cmidrule(lr){6-9}\cmidrule(lr){10-13}
& C\&W.(all) & C\&W.(tail) & AA.(all) & AA.(tail)
& C\&W.(all) & C\&W.(tail) & AA.(all) & AA.(tail)
& C\&W.(all) & C\&W.(tail) & AA.(all) & AA.(tail) \\
\midrule
\rowcolor{gray!15}{RoBal} &35.52\% &27.82\%&32.29\%&24.53\%&19.42\%&19.08\%&17.99\%&17.66\%& 12.98\% & 12.12\% & 12.06\% & 11.29\%\\
{BSL } &32.62\% &25.76\%&29.64\%&22.59\%&19.23\%&18.25\%&17.34\%&16.56\%& 15.05\% & 13.50\% & 13.30\% & 11.96\%\\
 \midrule
\rowcolor{gray!15}PGD-AT &31.18\% &17.95\%&28.90\%&15.48\%&\textbf{19.52\%}&17.40\%&\textbf{17.42\%}&15.53\%&14.38\% & 11.61\% &12.74\% & 10.18\% \\
 + UDR &28.93\% &15.89\%&26.95\%&13.80\%&17.47\%&15.60\%&15.83\%&14.12\%& 13.31\% & 10.60\% & 11.93\% & 9.49\%\\
\rowcolor{gray!15} + DAFA &31.14\% &20.97\%&28.73\%&18.64\%&18.54\%&17.69\%&16.76\%&16.08\%&14.13\% & 12.06\% & 12.92\% & 11.09\%\\
 + RobustLT &32.70\% &20.51\%&29.97\%&17.75\%&18.88\%&17.25\%&16.91\%&15.54\%& 15.03\% & 12.10\% & 13.21\% & 10.96\%\\
\rowcolor{gray!15} + Ours &\textbf{35.91\%} &\textbf{28.04\%}&\textbf{33.01\%}&\textbf{25.09\%}&19.39\%&\textbf{19.04\%}&17.25\%&\textbf{17.03\%}& \textbf{15.06\%} & \textbf{13.99\%}& \textbf{13.45\%} & \textbf{12.29\%} \\
\midrule
TRADES & 31.81\%&18.39\%&30.78\% & 17.29\%&20.84\%&18.35\%&19.40\%&17.14\%&13.81\% & 10.91\%&12.91\% & 10.18\% \\
\rowcolor{gray!15} + UDR & 30.92\%&17.54\%&29.54\%&16.15\%&18.98\%&16.83\%&17.45\%&15.50\%&12.23\%&9.95\%&11.09\%&9.08\%\\
 + DAFA &33.01\% &22.45\%&31.22\%&20.66\%&20.41\%&19.20\%&19.03\%&18.01\%&13.25\% & 11.15\%& 12.33\% & 10.34\%\\
\rowcolor{gray!15} + RobustLT & 32.69\%&19.98\%&31.22\%&18.44\%&21.64\%&19.61\%&\textbf{20.21\%}&18.31\%&14.19\% &11.45\% &\textbf{13.06\%} & 10.53\%\\
 + Ours & \textbf{34.69\%}&\textbf{24.99\%}&\textbf{32.24\%}&\textbf{22.65\%}&\textbf{21.79\%}&\textbf{20.26\%}&19.94\%&\textbf{18.71\%}& \textbf{14.28\%} & \textbf{12.76\%}& 12.92\% & \textbf{11.54\%} \\
\midrule
\rowcolor{gray!15} AWP & 30.15\%&15.81\%&29.15\%&14.78\%&22.06\%&19.05\%&20.67\%&18.01\%& 12.48\% & 9.41\%&11.23\% & 8.34\% \\
+ UDR &32.35\% &19.05\%&29.92\%&16.41\%&21.84\%&19.34\%&20.03\%&17.84\%& 10.46\% & 8.38\% & 9.80\% & 7.63\%\\
\rowcolor{gray!15} + DAFA &35.37\% &25.14\%&32.99\%&22.76\%&23.54\%&21.26\%&21.37\%&19.45\%& 16.54\% & 13.05\% & 11.67\% & 8.86\%\\
+ RobustLT &37.11\% &28.38\%&35.01\%&24.70\%&23.80\%&21.22\%&21.50\%&19.23\% &17.14\% & 13.41\% &15.46\% & 12.15\%\\
\rowcolor{gray!15}+ Ours &\textbf{40.35\%} &\textbf{33.32\%}&\textbf{37.37\%}&\textbf{30.55\%}&\textbf{24.44\%}&\textbf{24.10\%}&\textbf{22.02\%}&\textbf{21.83\%} & \textbf{18.33\%} & \textbf{16.95\%} &\textbf{ 16.24\%} & \textbf{14.90\%}\\
\bottomrule
\end{tabular}}
\end{table*}

\renewcommand{\arraystretch}{0.95}
\begin{table}[htpb]
\centering
\setlength{\tabcolsep}{3pt}
\caption{Ablation study of CGRm on CIFAR10-LT. ‘PCC’ denotes the balanced prior-calibrated classification term in Eq. (\ref{eq:balanced-ce}), where the feedback reweighting $w_y^{(t)}$ and class-wise factor $\beta_y^{(t)}$ are given in Eq. (\ref{eq18}), and $\lambda_m$ in Eq. (\ref{eq:objective}) indicates whether the confusion-geometry margin term is used.}
\label{tab:main-comparison}
\footnotesize
\begin{tabular}{llcccccccc}
\toprule
\multirow{2}{*}{PCC} & \multirow{2}{*}{$w_y^{(t)}$}& \multirow{2}{*}{$\beta_y^{(t)}$}& \multirow{2}{*}{$\lambda_m$}
& \multicolumn{4}{c}{CIFAR10-LT}\\
\cmidrule(lr){5-8}
& & & & Nat.(all) & Nat.(tail) & Rob.(all) & Rob.(tail)\\
\midrule
\rowcolor{gray!15}$\times$&$\checkmark$&$\checkmark$&$\checkmark$ &56.10\% & 45.56\% & 32.38\% & 20.49\% \\
$\checkmark$&$\times$&$\times$&$\times$&64.60\% & 58.79\% & 32.96\% & 22.31\%\\
\rowcolor{gray!15}$\checkmark$&$\times$&$\checkmark$&$\checkmark$ &68.27\% & 62.01\% & 34.78\% & 25.04\%\\
$\checkmark$&$\checkmark$&$\times$&$\checkmark$ &67.78\% & 61.51\% & 34.47\% & 24.96\%\\
\rowcolor{gray!15}$\checkmark$&$\checkmark$&$\checkmark$&$\times$ &67.92\% & 61.67\% & \textbf{35.41\%} & 24.61\%\\
\midrule
$\checkmark$&$\checkmark$&$\checkmark$&$\checkmark$&\textbf{68.29\%} &\textbf{62.39\%} &34.92\% & \textbf{27.06\%} \\
\bottomrule
\end{tabular}
\end{table}


\subsection{Experimental details}
\textbf{Datasets.} 
Following RobustLT, we apply the exponential long-tailed sampling strategy \citep{cao2019ldam} to derive CIFAR10‑LT, CIFAR100‑LT, and TinyImageNet‑LT (200 classes) from CIFAR10, CIFAR100 \cite{Krizhevsky09learningmultiple}, and TinyImageNet \cite{le2015tiny}, respectively. For \(C\) classes with imbalance ratio \(K\), the number of training samples for class \(i\) is
\begin{equation}
   n_i = n_{\max} K^{-\frac{i}{C-1}},
\end{equation}
where classes are sorted from head to tail. We set the imbalance ratio to \(50\) for CIFAR10-LT and \(10\) for CIFAR100-LT and TinyImageNet-LT. The test sets retain the original balanced class distributions.

\textbf{Compared Methods.}
We compare against advanced AT techniques, including conventional debiased AT methods (RoBal \citep{wu2021robal}, REAT \citep{li2023reat}, BSL \citep{yue2024revisiting}, TAET \citep{wang2025taet}) and plug‑in methods (UDR \citep{bui2022udr}, CFA \citep{wei2023cfa}, DAFA \citep{lee2024dafa}, RobustLT \citep{zhang2026robustlt}) that are applied on base AT algorithms such as PGD‑AT \citep{madry2018towards}, TRADES \citep{zhang2019trades}, and AWP \citep{wu2020awp}. For fair comparison, each plug-in method shares the same learner, backbone, and attack setting as its corresponding baseline.

\textbf{Evaluation Metrics.}
We report natural and robust accuracy on all test classes (Nat.(all) and Rob.(all)), as well as on the 80\% of classes with fewest training samples (Nat.(tail) and Rob.(tail)). Robust accuracy is measured under a 20-step \(\ell_\infty\) PGD attack with radius \(8/255\) and step size \(2/255\). As an additional sanity check, we also evaluate the final models with AutoAttack to rule out gradient masking.

\textbf{Implementation Details.} The main experiments use ResNet18, PreActResNet \cite{he2016identity}, WideResNet \cite{zagoruyko2016wide} as backbones, optimized with SGD (momentum 0.9, weight decay \(5\times10^{-4}\)) and an initial learning rate of 0.1. All models are trained for 110 epochs. The perturbation radius is set to \(\epsilon=8/255\). Adversarial examples are generated by 10-step PGD during training and evaluated by 20-step PGD at test time. The base AT configuration is kept unchanged.
For CGRm-specific settings, the robust-error feedback and the confusion-geometry graph are updated every \(10\) epochs. For each source class, the graph retains the top-\(3\) most relevant target classes. The resulting graph is used only in the margin regularization term with weight \(\lambda_{\mathrm{m}}=10/C\), where $C$ denotes the number of classes. All other CGRm-specific hyperparameters are fixed across datasets and listed in the appendix.

\subsection{Comparative Results}

Tables~\ref{tab1}-\ref{tab3} present a comprehensive comparison across datasets, backbones, and base learners. 
Compared with existing plug‑in methods (e.g., UDR, CFA, DAFA, and RobustLT), CGRm consistently enhances both clean and adversarial performance, with the most stable gains observed on tail classes. For example, under ResNet‑based PGD‑AT on CIFAR10‑LT, CGRm increases Nat.(tail)/Rob.(tail) from 46.81\%/18.60\% (RobustLT) to 66.15\%/27.93\%. On CIFAR100‑LT with TRADES, it improves the best competitor’s Rob.(all)/Rob.(tail) from 20.68\%/18.25\% to 21.54\%/20.73\%, while also raising Nat.(all) from 45.50\% to 48.81\%. These gains indicate that the proposed method  allocates robust learning pressure more effectively toward vulnerable classes and ambiguous boundaries.

The improvement persists when switching from ResNet to Pre‑ResNet and WideResNet. On Pre‑ResNet, CGRm attains strong tail robustness. In the sole case where RobustLT marginally surpasses CGRm in overall robustness—TinyImageNet‑LT with TRADES (14.08\% vs. 13.96\% Rob.(all))—CGRm still improves Rob.(tail) from 11.22\% to 13.28\%. On WideResNet, the benefits scale with model capacity: CGRm boosts PGD‑AT on CIFAR10‑LT from 31.54\%/19.11\% to 35.64\%/26.41\% in Rob.(all)/Rob.(tail), and AWP combined with CGRm reaches 43.72\%/37.79\% on CIFAR10‑LT and 26.06\%/25.85\% on CIFAR100‑LT. Collectively, these comparisons demonstrate that CGRm effectively addresses the core robustness failure induced by class imbalance.

Table~\ref{tab:attack-wrn} compares different methods under more stronger and diverse attacks, C\&W \cite{carlini2017towards} and AutoAttack \cite{croce2020reliable}. Similarly, CGRm improves tail robustness across base learners and datasets, while preserving competitive all-class robustness. Under PGD-AT, CGRm increases CIFAR10-LT C\&W.(tail)/AA.(tail) from 20.51\%/17.75\% with RobustLT to 28.04\%/25.09\%, and improves TinyImageNet-LT from 12.10\%/10.96\% to 13.99\%/12.29\%. Similar gains are observed with TRADES, where CGRm obtains stronger tail robustness than RobustLT on CIFAR10-LT (24.99\%/22.65\% vs. 19.98\%/18.44\%) and CIFAR100-LT (20.26\%/18.71\% vs. 19.61\%/18.31\%). With AWP, CGRm achieves the most consistent improvement, giving the best result on all reported metrics, including 40.35\%/33.32\% C\&W accuracy and 37.37\%/30.55\% AutoAttack accuracy on CIFAR10-LT in all/tail classes. 
We note that a few all-class entries remain close or slightly favor competing methods, such as CIFAR100-LT AA.(all) under TRADES. However, the corresponding tail metrics are consistently improved by CGRm.

\subsection{Ablation Study}
Table~\ref{tab:main-comparison} ablates the four components of CGRm: the prior-calibrated classification term (PCC) in Eq. (\ref{eq:balanced-ce}), feedback weight \(w_y\) and class-wise coefficient \(\beta_y\) in Eq. (\ref{eq18}), and confusion-geometry margin in Eq. (\ref{eq:objective}). 1) Removing PCC drops natural accuracy from 68.29\% to 56.10\% (all) and from 62.39\% to 45.56\% (tail), and tail robust accuracy from 27.06\% to 20.49\%. 2) The two feedback components are complementary. Removing \(w_y\) lowers Rob.(tail) to 25.04\%, indicating vulnerable source classes need stronger optimization contribution. Removing \(\beta_y\) gives a similar drop to 24.96\%, showing that loss reweighting alone is insufficient; consistency regularization must also be strengthened for under-robust classes. 3) Without the margin term, Rob.(all) slightly increases to 35.41\%, but Rob.(tail) falls to 24.61\%. Class-wise feedback alone improves average robustness, yet tail robustness requires correcting specific source-target confusions. The graph-guided margin thus trades a small average gain for substantial tail improvement. 4) Retaining PCC while disabling all three robust modules yields 64.60\% Nat.(all), 58.79\% Nat.(tail), 32.96\% Rob.(all), and 22.31\% Rob.(tail). Compared to this baseline, the full model improves all metrics, especially Rob.(tail) by 4.75 points. 

\section{Conclusion}

This paper presents CGRm, a confusion-geometry rebalanced framework for AT under long-tailed distributions. Our key insight is that robust imbalance cannot be adequately captured by class‑wise accuracy alone: adversarial errors tend to originate from vulnerable source classes and flow toward specific target classes, with the errors often arising from geometrically plausible tail‑to‑head confusions. To address this, CGRm converts the periodic robust evaluation into three components: source‑class loss weights, class‑wise robust coefficients, and a directed confusion‑geometry graph. By integrating feedback‑weighted robust optimization with graph‑guided margin correction, CGRm effectively strengthens under‑robust classes and hardens the decision boundaries most responsible for long‑tailed robust collapse. Extensive experiments and ablation studies demonstrate that our design improves robust performance over existing long‑tailed AT methods, confirming the effectiveness of both robust‑error feedback and confusion‑geometry boundary correction.

\bibliography{aaai2027}

@inproceedings{goodfellow2015explaining,
  title     = {Explaining and Harnessing Adversarial Examples},
  author    = {Goodfellow, Ian J. and Shlens, Jonathon and Szegedy, Christian},
  booktitle = {International Conference on Learning Representations},
  year      = {2015}
}

@inproceedings{madry2018towards,
  title     = {Towards Deep Learning Models Resistant to Adversarial Attacks},
  author    = {Madry, Aleksander and Makelov, Aleksandar and Schmidt, Ludwig and Tsipras, Dimitris and Vladu, Adrian},
  booktitle = {International Conference on Learning Representations},
  year      = {2018}
}

@TECHREPORT{Krizhevsky09learningmultiple,
  author = {Alex Krizhevsky and Geoffrey Hinton},
  title = {Learning multiple layers of features from tiny images},
  institution = {University of Toronto},
  year = {2009}
}

@article{le2015tiny,
  title = {Tiny imagenet visual recognition challenge},
  author = {Le, Yann and Yang, Xuan},
  journal = {CS 231N},
  volume = {7},
  number = {7},
  pages = {3},
  year = {2015}
}

@inproceedings{carlini2017towards,
  title     = {Towards Evaluating the Robustness of Neural Networks},
  author    = {Carlini, Nicholas and Wagner, David},
  booktitle = {2017 IEEE Symposium on Security and Privacy (SP)},
  pages     = {39--57},
  year      = {2017},
  publisher = {IEEE}
}

@inproceedings{croce2020reliable,
  title     = {Reliable evaluation of adversarial robustness with an ensemble of diverse parameter-free attacks},
  author    = {Croce, Francesco and Hein, Matthias},
  booktitle = {Proceedings of International Conference on Machine Learning (ICML)},
  pages     = {1126--1135},
  year      = {2020}
}

@inproceedings{zhang2019trades,
  title     = {Theoretically Principled Trade-off between Robustness and Accuracy},
  author    = {Zhang, Hongyang and Yu, Yaodong and Jiao, Jiantao and Xing, Eric P. and El Ghaoui, Laurent and Jordan, Michael I.},
  booktitle = {International Conference on Machine Learning},
  pages     = {7472--7482},
  year      = {2019}
}

@inproceedings{wu2020awp,
  title     = {Adversarial Weight Perturbation Helps Robust Generalization},
  author    = {Wu, Dongxian and Xia, Shu-Tao and Wang, Yisen},
  booktitle = {Advances in Neural Information Processing Systems},
  pages     = {2958--2969},
  year      = {2020}
}

@inproceedings{cao2019ldam,
  title     = {Learning Imbalanced Datasets with Label-Distribution-Aware Margin Loss},
  author    = {Cao, Kaidi and Wei, Colin and Gaidon, Adrien and Arechiga, Nikos and Ma, Tengyu},
  booktitle = {Advances in Neural Information Processing Systems},
  year      = {2019}
}

@inproceedings{yue2024revisiting,
  title={Revisiting adversarial training under long-tailed distributions},
  author={Yue, Xinli and Mou, Ningping and Wang, Qian and Zhao, Lingchen},
  booktitle={Proceedings of the IEEE/CVF conference on computer vision and pattern recognition},
  pages={24492--24501},
  year={2024}
}

@inproceedings{ren2020balanced,
  title     = {Balanced Meta-Softmax for Long-Tailed Visual Recognition},
  author    = {Ren, Jiawei and Yu, Cunjun and Ma, Xiao and Zhao, Haiyu and Yi, Shuai and others},
  booktitle = {Advances in Neural Information Processing Systems},
  pages     = {4175--4186},
  year      = {2020}
}

@inproceedings{wu2021robal,
  title     = {Adversarial Robustness under Long-Tailed Distribution},
  author    = {Wu, Tong and Liu, Ziwei and Huang, Qingqiu and Wang, Yu and Lin, Dahua},
  booktitle = {Proceedings of the IEEE/CVF Conference on Computer Vision and Pattern Recognition},
  pages     = {8659--8668},
  year      = {2021}
}

@inproceedings{bui2022udr,
  title     = {A Unified Wasserstein Distributional Robustness Framework for Adversarial Training},
  author    = {Bui, Anh Tuan and Le, Trung and Tran, Quan Hung and Zhao, He and Phung, Dinh},
  booktitle = {International Conference on Learning Representations},
  year      = {2022}
}

@inproceedings{wei2023cfa,
  title     = {CFA: Class-Wise Calibrated Fair Adversarial Training},
  author    = {Wei, Zeming and Wang, Yifei and Guo, Yiwen and Wang, Yisen},
  booktitle = {Proceedings of the IEEE/CVF Conference on Computer Vision and Pattern Recognition},
  pages     = {8193--8201},
  year      = {2023}
}

@inproceedings{lee2024dafa,
  title     = {DAFA: Distance-Aware Fair Adversarial Training},
  author    = {Lee, Hyungyu and Lee, Saehyung and Jang, Hyemi and Park, Junsung and Bae, Ho and Yoon, Sungroh},
  booktitle = {International Conference on Learning Representations},
  year      = {2024}
}

@misc{li2023reat,
  title         = {Alleviating the Effect of Data Imbalance on Adversarial Training},
  author        = {Li, Guanlin and Xu, Guowen and Zhang, Tianwei},
  year          = {2023},
  eprint        = {2307.10205},
  archivePrefix = {arXiv},
  primaryClass  = {cs.LG}
}

@inproceedings{wang2025taet,
  title     = {TAET: Two-Stage Adversarial Equalization Training on Long-Tailed Distributions},
  author    = {Wang, Yu-Hang and Guo, Junkang and Liu, Aolei and Wang, Kaihao and Wu, Zaitong and Liu, Zhenyu and Yin, Wenfei and Liu, Jian},
  booktitle = {Proceedings of the IEEE/CVF Conference on Computer Vision and Pattern Recognition},
  pages     = {15476--15485},
  year      = {2025}
}

@misc{zhao2024atsurvey,
  title         = {Adversarial Training: A Survey},
  author        = {Zhao, Mengnan and Zhang, Lihe and Ye, Jingwen and Lu, Huchuan and Yin, Baocai and Wang, Xinchao},
  year          = {2024},
  eprint        = {2410.15042},
  archivePrefix = {arXiv},
  primaryClass  = {cs.LG},
  doi           = {10.48550/arXiv.2410.15042}
}

@inproceedings{zhao2023smooth,
  title     = {Fast Adversarial Training with Smooth Convergence},
  author    = {Zhao, Mengnan and Zhang, Lihe and Kong, Yuqiu and Yin, Baocai},
  booktitle = {Proceedings of the IEEE/CVF International Conference on Computer Vision},
  pages     = {4720--4729},
  year      = {2023}
}

@misc{zhang2026robustlt,
  title         = {Taming the Long Tail: Rebalancing Adversarial Training via Adaptive Perturbation},
  author        = {Zhang, Lilin and Guo, Yimo and Li, Yue and Shi, Jiancheng and Liu, Xianggen},
  year          = {2026},
  eprint        = {2506.13395},
  archivePrefix = {arXiv},
  primaryClass  = {cs.LG}
}

@article{du2024probabilistic,
  title={Probabilistic contrastive learning for long-tailed visual recognition},
  author={Du, Chaoqun and Wang, Yulin and Song, Shiji and Huang, Gao},
  journal={IEEE Transactions on Pattern Analysis and Machine Intelligence},
  volume={46},
  number={9},
  pages={5890--5904},
  year={2024},
  publisher={IEEE}
}

@inproceedings{robey2024adversarial,
  title={Adversarial training should be cast as a non-zero-sum game},
  author={Robey, Alex and Latorre, Fabian and Pappas, George and Hassani, Hamed and Cevher, Volkan},
  booktitle={International Conference on Learning Representations},
  volume={2024},
  pages={57834--57854},
  year={2024}
}

@article{yue2023revisiting,
  title={Revisiting adversarial robustness distillation from the perspective of robust fairness},
  author={Yue, Xinli and Ningping, Mou and Wang, Qian and Zhao, Lingchen},
  journal={Advances in Neural Information Processing Systems},
  volume={36},
  pages={30390--30401},
  year={2023}
}

@article{jia2024improving,
  title={Improving fast adversarial training with prior-guided knowledge},
  author={Jia, Xiaojun and Zhang, Yong and Wei, Xingxing and Wu, Baoyuan and Ma, Ke and Wang, Jue and Cao, Xiaochun},
  journal={IEEE Transactions on Pattern Analysis and Machine Intelligence},
  volume={46},
  number={9},
  pages={6367--6383},
  year={2024},
  publisher={IEEE}
}

@article{jiang2026rethinking,
  title={Rethinking Frequency Modeling: Tail-Aware Dynamic Adversarial Training for Long-Tailed Robustness},
  author={Jiang, Chengze and Dong, Minjing and Wang, Zhuangzhuang and Gui, Jie and Jia, Ju and Tang, Yuan Yan and Kwok, James Tin-Yau},
  journal={IEEE Transactions on Information Forensics and Security},
  year={2026},
  publisher={IEEE}
}

@inproceedings{zhao2024coblessing,
  title     = {Catastrophic Overfitting: A Potential Blessing in Disguise},
  author    = {Zhao, Mengnan and Zhang, Lihe and Kong, Yuqiu and Yin, Baocai},
  booktitle = {European Conference on Computer Vision},
  pages     = {293--310},
  year      = {2024}
}

@inproceedings{zhao2026erroramplification,
  title     = {Mitigating Error Amplification in Fast Adversarial Training},
  author    = {Zhao, Mengnan and Zhang, Lihe and Wang, Bo and Zheng, Tianhang and Zhong, Hong and Min, Geyong},
  booktitle = {Proceedings of the IEEE/CVF Conference on Computer Vision and Pattern Recognition},
  year      = {2026}
}

@inproceedings{qin2026fedcart,
  title={FedCART: Tackling Long-Tailed Distributions in Federated Adversarial Training via Classifier Refinement},
  author={Qin, Yuchen and Zhou, Yizhi and Wang, Junxiao and Xie, Xin and Qi, Heng},
  booktitle={Proceedings of the IEEE/CVF Conference on Computer Vision and Pattern Recognition},
  pages={24557--24566},
  year={2026}
}

@inproceedings{li2025head,
  title={From head to tail: efficient black-box model inversion attack via long-tailed learning},
  author={Li, Ziang and Zhang, Hongguang and Wang, Juan and Chen, Meihui and Hu, Hongxin and Yi, Wenzhe and Xu, Xiaoyang and Yang, Mengda and Ma, Chenjun},
  booktitle={Proceedings of the Computer Vision and Pattern Recognition Conference},
  pages={29288--29298},
  year={2025}
}

@article{zagoruyko2016wide,
  title={Wide residual networks},
  author={Zagoruyko, Sergey and Komodakis, Nikos},
  journal={arXiv preprint arXiv:1605.07146},
  year={2016}
}

@inproceedings{he2016identity,
  title={Identity mappings in deep residual networks},
  author={He, Kaiming and Zhang, Xiangyu and Ren, Shaoqing and Sun, Jian},
  booktitle={European conference on computer vision},
  pages={630--645},
  year={2016},
  organization={Springer}
}


\end{document}